\documentclass[letterpaper]{article} 
\usepackage[preprint]{aaai2027}
\usepackage[hyphens]{url}  
\usepackage{graphicx} 
\usepackage{natbib}  
\usepackage{caption} 
\usepackage{algorithm}
\usepackage{algorithmic}
\usepackage{amsmath}
\usepackage{amsfonts}
\usepackage{amssymb}

\usepackage{newfloat}
\usepackage{listings}
\DeclareCaptionStyle{ruled}{labelfont=normalfont,labelsep=colon,strut=off} 
\floatstyle{ruled}
\newfloat{listing}{tb}{lst}{}
\floatname{listing}{Listing}

\usepackage{booktabs}
\usepackage{multirow}
\usepackage{graphicx}
\title{ForeSeek: Foresight-Driven Reasoning for Long-Form Video Understanding}
\author{
    Yuchen Sun\textsuperscript{\rm 1}\equalcontrib,
    Siyuan Cheng\textsuperscript{\rm 1}\equalcontrib,
    Shi-Zhe Chen\textsuperscript{\rm 1}\corresponding\thanks{Project lead.},
    Yixin Qin\textsuperscript{\rm 1},
    Jinwen Luo\textsuperscript{\rm 2},
    Zheng Wei\textsuperscript{\rm 1}\corresponding
}
\affiliations{
    \textsuperscript{\rm 1}Tencent PCG \quad
    \textsuperscript{\rm 2}Tencent CSIG\\
    {\small \texttt{\{harryycsun, chancecheng, shizhechen, wadeqin, jamsluo, hemingwei\}@tencent.com}}
}

\providecommand{\our}{ForeSeek}

\begin{document}

\maketitle

\begin{abstract}
Long-form video understanding, as a fundamental yet crucial vision task, is being revolutionized by multimodal large language model (MLLM)-based video agents. However, the retrieval-driven nature inherent to these agents creates a significant bottleneck, forcing them to reactively fetch clips based on local saliency without a global temporal plan. To address this, we present \textbf{ForeSeek}, a foresight-driven video agent that transforms long-video reasoning from reactive retrieval into predictive exploration. Rather than blindly probing the timeline, ForeSeek forms visually grounded expectations over unseen temporal regions, verifies them through targeted observations, and continuously refines its search based on emerging evidence, where failed expectations serve as informative signals for principled backtracking. Furthermore, we develop a two-stage training framework that equips MLLMs with foresight-guided reasoning and efficient decision making. Extensive experiments on five long-video benchmarks demonstrate that ForeSeek consistently improves the efficiency--accuracy trade-off of long-video reasoning, achieving competitive or superior accuracy while observing $2.5$--$4.4\times$ fewer frames than the strongest search-based agent.
Code is available at \url{https://github.com/harry0021/ForeSeek}.
\end{abstract}

\section{Introduction}

Multimodal Large Language Models (MLLMs) have achieved remarkable progress in video understanding~\cite{GPT-5, Gemini2.5, Qwen3.5-Omni}.
However, long-form video understanding remains challenging, as videos often contain thousands of redundant frames with long-range temporal dependencies, making uniform frame sampling computationally prohibitive and prone to irrelevant noise that degrades reasoning~\cite{SeViLA, AdaFrame}.
Consequently, effective long-video understanding is becoming less about recognizing individual frames and more about deciding which observations are worth acquiring.

\begin{figure}[!t]
    \centering
    \includegraphics[width=0.495\textwidth]{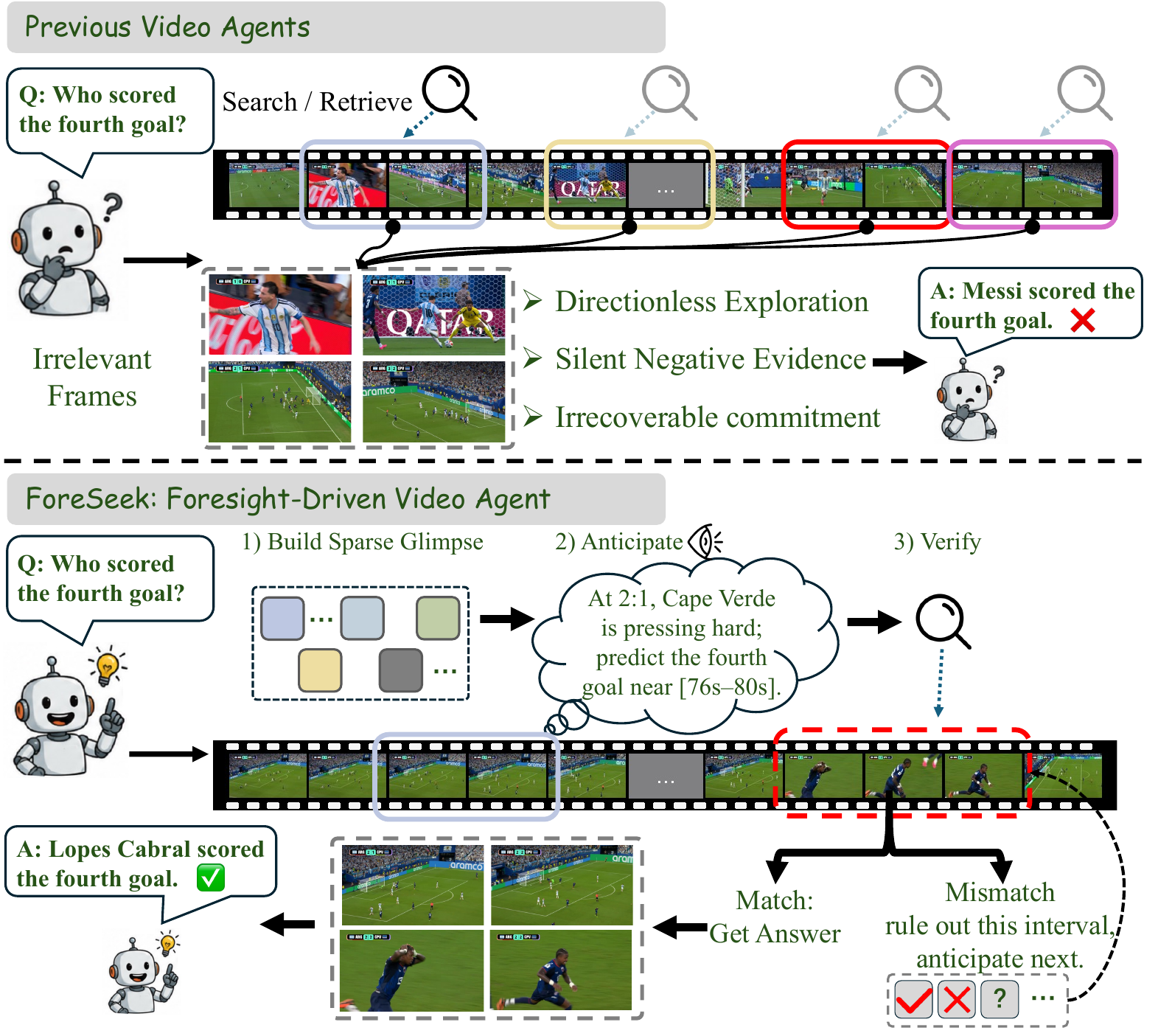}
    \caption{
    \textbf{Prior Agents vs.\ \our{}.}
    Prior agents \textbf{(top)} reactively probe the video timeline without explicit expectations about unseen content, 
    whereas ~\our{} \textbf{(bottom)} performs foresight-guided reasoning by maintaining predictive expectations that continuously guide evidence collection.
    }
    \label{fig:intro}
\end{figure}

To overcome the limitations of passive, single-pass video understanding, the video agent paradigm has recently emerged as a promising direction~\cite{VideoAgent, LongVideo-R1}.
Rather than relying on a single forward pass, a video agent iteratively invokes tools, samples frames, and aggregates evidence.
Existing efforts make this loop effective through three main strategies: pre-captioning or indexing videos for iterative clip re-ranking~\cite{VideoAgent, VideoAgent2}, adaptively invoking multi-granularity tools to determine where and how densely to sample~\cite{VideoSeek, EVA}, and learning frame-selection policies end-to-end through reinforcement learning~\cite{FrameThinker, TimeSearch-R}.
Existing video agents have substantially advanced long-video understanding by enabling MLLMs to iteratively retrieve and inspect informative clips. Despite their diverse designs, these approaches largely follow a retrieval-driven reasoning paradigm, where progress is made through continuously acquiring additional observations. Consequently, agents remain inherently reactive, searching for locally relevant evidence without forming explicit expectations about the evolution of unseen content.
This reactive paradigm introduces three fundamental limitations:
\emph{(i) Directionless exploration:} frame selection is guided by local relevance rather than global temporal anticipation, leading to inefficient exploration.
\emph{(ii) Unexploited negative evidence:} failed retrievals are treated as discarded attempts rather than informative signals that constrain future search.
\emph{(iii) Irreversible commitment:} once an incorrect temporal hypothesis is pursued, agents lack a mechanism to revise their beliefs and redirect exploration at a global level.
As a result, frame consumption scales poorly: agents either overspend computation on exhaustive search or miss critical moments required for accurate reasoning.

To address this, we draw inspiration from human predictive perception~\cite{clark2013whatever}. 
Humans do not perceive long temporal sequences by exhaustive observation; instead, they form predictions from sparse evidence and continuously verify or revise them as new observations arrive.
Crucially, such anticipation is grounded in visual evidence, making it falsifiable rather than speculative.
This predictive capability allows humans to narrow the space of plausible future events before committing additional attention, thereby reducing unnecessary observations.
Motivated by this observation, we introduce a \textbf{FOR}esight-driven \textbf{E}vidence \textbf{Seek}ing video agent (\textbf{\our}) that enables MLLMs with predictive temporal reasoning (Fig.~\ref{fig:intro}).
Instead of treating retrieval as the reasoning process itself, \our\ maintains explicit expectations over unseen temporal regions and continuously refines these expectations as new evidence becomes available. 
Consequently, observations serve not merely as retrieved evidence but as tests of the agent's evolving hypotheses, allowing both supporting and contradicting evidence to guide future exploration.
In this way, reasoning is driven not by accumulating observations alone, but by progressively reducing uncertainty over the unseen video.
We further propose a two-stage learning framework to explicitly supervise foresight reasoning, rendering predictive reasoning a core capability instead of an implicit frame retrieval byproduct.

Extensive experiments across five long-video benchmarks demonstrate that \our\ consistently achieves a strong efficiency--accuracy trade-off among comparable video agents. 
It maintains competitive or superior reasoning performance under a compact frame budget, validating the effectiveness of foresight-guided exploration.

Our main contributions are summarized as follows:
\begin{itemize}

\item We introduce \textbf{\our}, a foresight-driven video agent that shifts long-video reasoning from reactive retrieval to predictive exploration by maintaining visually grounded, falsifiable expectations over unseen temporal regions to guide evidence collection.

\item We develop a global interval belief mechanism that iteratively verifies and updates predictions over unseen regions, turning failed predictions into negative evidence that deprioritizes implausible regions and redirects exploration globally. We further enable foresight through a two-stage training framework that learns from structured predictive trajectories and an efficiency-aware trajectory-level reward.

\item Extensive experiments on five long-video benchmarks demonstrate that \our\ achieves a strong efficiency–accuracy trade-off among open-source video agents, delivering competitive or superior accuracy with a compact frame budget.

\end{itemize}

\section{Related Work}

\subsection{Long-Form Video Understanding}
Multimodal Large Language Models (MLLMs)~\cite{Gemini2.5, Qwen3.5-Omni, GPT-5, LLaVA1_5} have rapidly extended their capabilities from static images to videos~\cite{InternVideo2_5, InternVL3.5, VideoLLaMA3, MiniGPT4-Video, AVS_SVC, ReMoRa, LF-VILA}. 
Still, long-form video remains particularly challenging: redundant frames, long-range temporal dependencies, and limited context windows lead to both prohibitive cost and degraded reasoning when feeding densely or uniformly sampled frames~\cite{SeViLA, AdaFrame, longvideobench, lvbench}. 
To mitigate this, prior work has explored token compression and memory aggregation~\cite{MovieChat, LongVLM}, hierarchical or chunk-wise summarization~\cite{VideoTree, LLoVi}, and other single-pass strategies that consume the visual content once without revisiting specific moments under the query.

\subsection{Video Agents and RL-based Video Reasoning}
Video agents formulate long-video understanding as an iterative tool-using process, where MLLMs progressively gather evidence through external actions~\cite{VideoAgent,EVA,VideoSeek,Conan,VideoMind,R-VLM}.
One line of work focuses on designing the search framework itself: VideoAgent~\cite{VideoAgent} pioneers iterative LLM-guided retrieval and VideoAgent2~\cite{VideoAgent2} refines it with uncertainty- and confidence-aware reasoning, while VideoSeek~\cite{VideoSeek} adopts a think--act--observe framework with a multi-granularity toolkit and EVA~\cite{EVA} advocates a planning-before-perception paradigm.
A complementary line explicitly learns retrieval policies via reinforcement learning: FrameThinker~\cite{FrameThinker} formulates frame selection as a GRPO-trained decision process, TimeSearch-R~\cite{TimeSearch-R} adds completeness-aware temporal search, Video-R1~\cite{Video-R1} performs trajectory-level RL, and EvoVid~\cite{EvoVid} exploits questioner--solver self-play.
Despite different objectives, these methods all remain retrieval-centric, predicting \emph{where to observe next} rather than \emph{what to expect in unseen video}. VideoHV-Agent~\cite{VideoHV-Agent} moves closer to predictive reasoning by verifying hypotheses before retrieval, but its hypotheses target candidate \emph{answers} rather than future video content, without a global predictive representation of the timeline.
In contrast, \our{} formulates visually grounded hypotheses over unseen video and uses verification outcomes to confirm, refine, or reject them, so failed predictions demote implausible regions and redirect exploration at a global level.

\subsection{Anticipation and Predictive Coding in Vision}
Predictive perception---the view that the brain continuously projects expectations and explains away sensory input via prediction errors---is a long-standing theme in cognitive science~\cite{clark2013whatever} that has inspired vision work on action anticipation, future-frame prediction, and anticipatory captioning~\cite{AVT, Causality}. 
To our knowledge, however, this perspective has not been instantiated as the \emph{search target} of long-video agents. 
\our\ fills this gap with a foresight-driven paradigm where visually grounded forecasts over unseen intervals guide search through verification feedback.

\section{Method}
\paragraph{Task Definition.}
Given a long-form video $V$ of duration $T$, a question $q$, candidate answers
$\mathcal{A}=\{a_1,\dots,a_K\}$, and a visual-token budget $\mathcal{V}$, the
goal is to select the correct answer through adaptive visual inspection:
$\hat{a}=\operatorname{VideoAgent}_{\theta}(V,q;\mathcal{V})\in\mathcal{A}.$
Unlike uniform video consumption, the agent adaptively selects which regions
to observe and terminates once sufficient evidence is collected.

\begin{figure*}[!t]
\centering 
\includegraphics[width=0.90\textwidth]{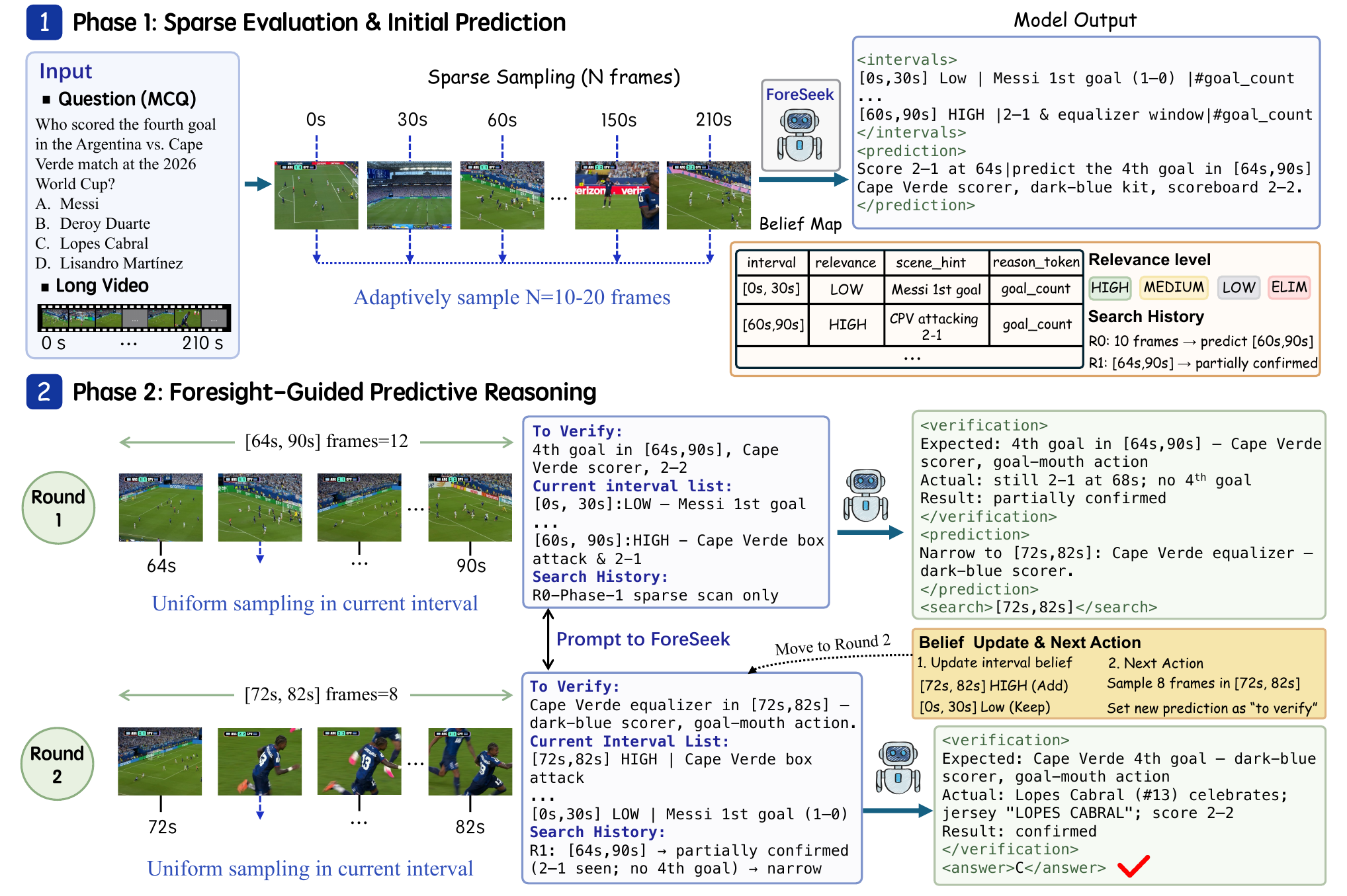}
\caption{
\textbf{Overview of \our.}
\our\ enables \emph{foresight-guided reasoning} by anticipating unseen regions and refining beliefs through evidence.
\textbf{Phase 1:} sparse glimpses initialize a temporal belief and prediction.
\textbf{Phase 2:} verification updates beliefs, with failed predictions guiding search redirection.
}
\label{fig:main}
\end{figure*}

\subsection{\our: Foresight-Guided Reasoning}
Unlike existing video agents that reason via incremental observation retrieval, \our\ explicitly sustains a \emph{predictive state} throughout long-video understanding: the agent builds expectations for unseen temporal regions and uses newly obtained evidence to verify, adjust, or discard them (Fig.~\ref{fig:main}), so retrieval becomes a means of testing the current prediction rather than the primary driver of reasoning.
Formally, at reasoning step $r$ the predictive state is a triple
$\mathcal{S}_r=(\mathcal{B}_r,\phi_r,\mathcal{H}_{<r})$,
where $\mathcal{B}_r$ is a global interval-level \textbf{belief} over the timeline of $V$,
$\phi_r$ is a visually grounded \textbf{expectation} of what an unseen target interval should contain,
and $\mathcal{H}_{<r}$ is the reasoning trace up to step $r$.

\paragraph{Predictive State Initialization.}
Foresight-guided reasoning initiates with a sparse, query-aware scan over the whole video. Probing budget allocation follows video duration:
\begin{equation}
\label{eq:phase1_sampling}
N(T)=\mathrm{clip}\!\Big(
\big\lfloor N_{\min}+\alpha\log_2\!\big(1+\tfrac{T}{D_0}\big)\big\rceil,
N_{\min},N_{\max}
\Big),
\end{equation}
where $[N_{\min},N_{\max}]=[10,20]$, $D_0=600\,\mathrm{s}$, and $\alpha=3$.
The duration-adaptive nonlinear scan provides adequate coverage for short
videos while avoiding excessive sampling of long ones.
Given the sampled frames and $q$, $\pi_\theta$ produces the initial belief
$\mathcal{B}_1$, expectation $\phi_1$.

\paragraph{Evidence-driven Predictive Reasoning.}
Starting from $\mathcal{S}_1$, \our\ progressively refines its understanding
through targeted evidence collection.
At step $r$, the agent samples $n_r\in[4,16]$ frames from the current search
interval $[t_r^{\mathrm{s}},t_r^{\mathrm{e}}]$ and prompts $\pi_\theta$ with
$\mathcal{S}_r$.
The model verifies whether the observed evidence supports $\phi_r$, yielding
$v_r\in\{\textit{confirmed},\textit{partially confirmed},\textit{not found}\}$:
\emph{(i) Confirmed:} the evidence supports $\phi_r$ and the agent may answer.
\emph{(ii) Partially confirmed:} $\phi_r$ is refined and a narrower sub-region
is proposed for the next round.
\emph{(iii) Not found:} the current prediction is rejected, its interval is
deprioritized, and the search is redirected to the most relevant remaining region.
Thus, negative evidence progressively shifts the search away from
unsuccessful regions.
\paragraph{Global Interval-level Belief.}
Throughout reasoning, \our\ maintains a global interval-level belief,
\begin{equation}
\mathcal{B}_r=
\{(\text{start}_i,\text{end}_i,\text{relevance}_i,\text{hint}_i)\}_{i=1}^{M_r},
\end{equation}
where each interval has a textual hint and a relevance label in
$\{\textsc{High},\textsc{Medium},\textsc{Low},\textsc{Elim}\}$.
\textsc{Medium} is the default, no-signal state; \textsc{High} indicates
positive evidence, \textsc{Low} a weak prior or failed search, and
\textsc{Elim} a model-declared hard exclusion after verification.
The scheduler prioritizes \textsc{High} over \textsc{Medium}, excludes
\textsc{Low}/\textsc{Elim} from normal selection, and uses \textsc{Low} only
as a fallback. After an unsuccessful search, the framework demotes a
\textsc{High}/\textsc{Medium} interval to \textsc{Low}, but never assigns
\textsc{Elim}; maintaining that state is the model's responsibility.
\begin{algorithm}[t]
\caption{\our\ training-data pipeline.}
\label{alg:data}
\small
\begin{algorithmic}[1]
\REQUIRE Raw QA pool $\mathcal{D}_{raw}$; answerer models $M_A$ and $M_B$; judge $M_J$; teacher $M_T$; SFT-initialized policy $\pi_{\mathrm{sft}}$
\ENSURE SFT dataset $\mathcal{D}_{sft}$, RL dataset $\mathcal{D}_{rl}$

\STATE  \hrulefill\ \textbf{Stage 1: QA Supplementation} \hrulefill
\FORALL{$(V,q)\in\mathcal{D}_{raw}$}
    \STATE Generate candidate answers with $M_A$ and $M_B$
    \IF{predictions agree}
        \STATE Accept the shared answer
    \ELSE
        \STATE Resolve disagreement with judge model $M_J$
    \ENDIF
    \STATE Add the verified tuple $(V,q,a,c)$ to $\mathcal{D}_{qa}$ ($a$: answer, $c$: rationale)
\ENDFOR

\STATE \hrulefill\ \textbf{Stage 2: Split and SFT Trajectory Synthesis}\hrulefill
\STATE Split $\mathcal{D}_{qa}$ evenly into pre-SFT/pre-RL pools $\mathcal{D}_{\mathrm{sft}}^{0}$ and $\mathcal{D}_{\mathrm{rl}}^{0}$
\FORALL{$(V,q,a,c)\in\mathcal{D}_{\mathrm{sft}}^{0}$}
    \STATE Run $M_T$ in deployment-faithful Phase-1 mode to obtain the initial predictive state
    \STATE Run $M_T$ in oracle mode to synthesize an evidence-grounded search trajectory
    \STATE Compile into a multi-round SFT conversation; add to $\mathcal{D}_{sft}$ if valid (discard failures)
\ENDFOR

\STATE \hrulefill\ \textbf{Stage 3: Pass@$K$ Filtering of RL Split} \hrulefill
\FORALL{$(V,q,a)\in\mathcal{D}_{\mathrm{rl}}^{0}$}
    \STATE Estimate pass@$K$ ($K{=}8$) with $\pi_{\mathrm{sft}}$ at temperature $0.7$
    \IF{$k_{\min}\le \mathrm{pass@}K \le k_{\max}$ ($k_{\min}{=}1, k_{\max}{=}6$)}
        \STATE Add $(V,q,a)$ to $\mathcal{D}_{rl}$
    \ENDIF
\ENDFOR

\RETURN $\mathcal{D}_{sft},\mathcal{D}_{rl}$
\end{algorithmic}
\end{algorithm}
\subsection{Foresight Trajectory Dataset Construction}
Because \our{} requires supervision for expectation generation, evidence verification, and belief refinement rather than final answers alone, we build a foresight-trajectory corpus via the three-stage pipeline of Alg.~\ref{alg:data}, drawing questions from CG-Bench~\cite{cgbench}, LongVideoBench~\cite{longvideobench}, VideoHolmes~\cite{videoholmes}, and the academic and YouTube subsets of LLaVA-Video~\cite{LLaVAVideo}.
We remove potential evaluation contamination by matching question IDs, normalized question text, and source-video identity.
As LongVideoBench withholds ground-truth answers, we annotate it with a dual-answerer-plus-judge protocol~\cite{LongViTU} (answerers GPT-5.2 and Gemini-3.5-Flash, judge Gemini-3.1-Pro); other sources keep their original annotations.

We split $\mathcal{D}_{qa}$ evenly into pre-SFT and pre-RL pools of $10{,}021$ questions each.
On each pre-SFT question the teacher $M_T$ (Gemini-3-Flash) makes two calls: a \emph{Phase-1} call restricted to sparse frames, matching inference-time information, and an \emph{oracle} call that also sees the ground truth and synthesizes the full reasoning trajectory.
The oracle must ground every step in observed evidence rather than directly reveal the answer, and failed predictions must exhibit the recovery pattern ``wrong region $\rightarrow$ \texttt{not\_found} $\rightarrow$ redirect $\rightarrow$ confirm.''
Discarding failed syntheses leaves $8{,}165$ unique SFT pairs (CG-Bench $35.9\%$, LongVideoBench $28.2\%$, LLaVA-Video Academic $17.0\%$ and YouTube $15.9\%$, VideoHolmes $3.0\%$).

For RL, we sample $8$ rollouts per pre-RL question from $\pi_{\mathrm{sft}}$ at temperature $0.7$ and keep the $4{,}663$ ($46.5\%$) answered correctly in $1$--$6$ trials (CG-Bench $48.1\%$, LLaVA-Video YouTube $23.1\%$ and Academic $17.0\%$, VideoHolmes $6.7\%$, LongVideoBench $5.1\%$).

\subsection{Two-Stage Training: SFT + GSPO}
\label{sec:training}
We train \our\ via two-stage optimization: supervised learning on foresight-guided trajectories followed by trajectory-level reinforcement learning for accuracy and reasoning efficiency.
Let $\pi_\theta$ denote the MLLM-based agent policy and $\tau=\{(\mathcal{S}_r,o_r)\}_{r=1}^{n}$ a rollout of $n$ reasoning rounds, where $o_r$ is the tokenized MLLM output at step $r$ (belief update, expectation, verdict, and optionally the final answer).

\paragraph{Stage 1: Cold-start Supervised Fine-tuning.}
We first initialize $\pi_\theta$ by supervised fine-tuning on the synthesized foresight trajectories $\mathcal{D}_{sft}$.
Unlike answer-only supervision, each trajectory reveals the full predictive reasoning process, including belief updates, expectation generation, and evidence verification.
We optimize the standard token-level negative log-likelihood over the assistant turns of every trajectory:
\begin{equation}
\mathcal{L}_{\mathrm{SFT}}(\theta)
=-\!\!\!\sum_{\tau\in\mathcal{D}_{sft}}\sum_{r=1}^{|\tau|}
\log\pi_\theta\!\big(o_r\,\big|\,\mathcal{S}_r,\,o_{<r}\big).
\end{equation}
This stage teaches the model to produce consistent foresight-driven trajectories and provides a strong initialization for the subsequent reinforcement stage.

\paragraph{Stage 2: Efficiency-aware GSPO.}
Starting from the SFT checkpoint $\pi_{\mathrm{sft}}$, we further optimize the policy with Group Sequence Policy Optimization (GSPO)~\cite{GSPO}.
While Stage 1 teaches the reasoning schema, SFT alone favors longer traces as additional observations often improve accuracy, leading to unnecessary frame consumption. We therefore introduce a length-dependent decay to make efficiency an explicit training signal:
\begin{equation}
R(\tau)=R_{\mathrm{ans}}(\tau)\cdot T(n),
\end{equation}
where $R_{\mathrm{ans}}(\tau)\in\{0,1\}$ indicates final-answer correctness, $n$ is the number of reasoning rounds actually used by $\tau$, and $T(n)$ is a piecewise round-count decay factor,
\begin{equation}
T(n)=
\begin{cases}
1.0, & n<4,\\
1 - 0.1\,(n-3), & 4 \le n \le \mathcal{R}_{\max}.
\end{cases}
\end{equation}
Trajectories that terminate within three rounds receive the full reward ($1.0$), while longer trajectories are linearly discounted down to $T(5) = 0.8$ at the round limit ($\mathcal{R}_{\max} = 5$). Since $T(n) \ge 0.8 > 0$ for all valid $n$, the reward remains strictly positive for any correct trajectory.
Crucially, $R_{\mathrm{ans}}(\tau)\cdot T(n)$ is multiplicative, not additive: incorrect trajectories receive zero reward regardless of length, rewarding efficiency only when correct and preventing premature termination.
Given $G$ rollouts $\{\tau^{(g)}\}_{g=1}^{G}$ for $(V,q)$, GSPO estimates
$\hat{A}^{(g)}=(R(\tau^{(g)})-\operatorname{mean}_g R)/
(\operatorname{std}_g R+\epsilon)$, where $\epsilon=10^{-8}$ and
$\hat{A}^{(g)}=0$ for zero-variance groups.
We optimize $\pi_\theta$ with the sequence-level GSPO objective, using asymmetric
clipping $[1-\epsilon_{\mathrm{low}},1+\epsilon_{\mathrm{high}}]$
($\epsilon_{\mathrm{low}} {=} 10^{-3}$, $\epsilon_{\mathrm{high}} {=} 1.5\times10^{-3}$)
and a loss-level $k_1$-KL penalty ($\beta=0.02$) against $\pi_{\mathrm{sft}}$.
With the efficiency-aware reward, this objective encourages termination upon sufficient evidence, improving the accuracy--efficiency trade-off over SFT.

\section{Experiment}

\subsection{Experimental setting}

\paragraph{Benchmarks.}
We evaluate on five long-video understanding benchmarks that jointly span diverse domains, long-form comprehension, and multi-step reasoning: \textbf{VideoMME}~\cite{videomme}, \textbf{LongVideoBench}~\cite{longvideobench}, \textbf{LVBench}~\cite{lvbench}, \textbf{VideoHolmes}~\cite{videoholmes}, and \textbf{MLVU}~\cite{MLVU}. All evaluations are conducted without subtitles.

\paragraph{Baselines.}
Following the taxonomy in Related Work, we compare against the two families of video agents built on comparable $\sim$7--8B open-source backbones. \emph{Search-framework agents}---which primarily innovate on the iterative retrieval pipeline---are represented by \textbf{EVA}~\cite{EVA} and \textbf{Conan}~\cite{Conan}. \emph{RL-trained retrieval agents}---which explicitly optimize retrieval policies with reinforcement learning---are represented by \textbf{VideoChat-R1}~\cite{VideoChat-R1}, \textbf{Video-R1}~\cite{Video-R1}, \textbf{Video-MTR}~\cite{Video-MTR}, and \textbf{FrameThinker}~\cite{FrameThinker}. As reference points, we additionally report proprietary or much larger multimodal models (GPT-4o, GPT-5, Gemini-3.1-Flash-Lite, Qwen2.5-VL-72B) and agents built on such backbones (VideoSeek~\cite{VideoSeek}, VideoAgent~\cite{VideoAgent}); these are not directly comparable and are shown only to contextualize the achievable frontier. 
Baseline results are taken from the original papers when available, with each
method's default frame budget listed in Table~\ref{tab:main_results} for
accuracy--efficiency comparison. ``Comparable'' denotes backbone scale, not
generation: most prior agents use Qwen2.5-VL-7B, whereas ForeSeek uses
Qwen3-VL-8B. We therefore measure method-level gains against our backbone
under uniform 32-frame sampling and treat cross-backbone comparisons only as
system-level evidence.

\paragraph{Implementation Details.}
We adopt \textbf{Qwen3-VL-8B}~\cite{Qwen3VL} as the foundation model and fine-tune it on $ 8\times$ H20 GPUs with full-parameter updates in both stages. 
Stage~1 runs SFT on $8$K foresight trajectories for one epoch (lr $2\times10^{-6}$) via LLaMA-Factory~\cite{llamafactory}; 
Stage~2 optimizes the policy with GSPO~\cite{GSPO} on $\sim4.7$K samples (lr $1\times10^{-6}$, rollout group $8$) via VeRL~\cite{verl} for one epoch with early stopping. 
All raw frames are resized to $0.8\times$ resolution during training and inference.

\begin{table*}[t]
\centering
\resizebox{\textwidth}{!}{%
\begin{tabular}{l c c c c c c c c c c}
\toprule
\multirow{2}{*}{\textbf{Method}} & 
\multicolumn{2}{c}{\textbf{VideoMME (w/o sub)}} & 
\multicolumn{2}{c}{\textbf{LongVideoBench}} &
\multicolumn{2}{c}{\textbf{LVBench (w/o sub)}} & 
\multicolumn{2}{c}{\textbf{VideoHolmes}} &
\multicolumn{2}{c}{\textbf{MLVU (w/o sub)}}\\
\cmidrule(lr){2-3} \cmidrule(lr){4-5} \cmidrule(lr){6-7} \cmidrule(lr){8-9} \cmidrule(lr){10-11}
& \#Frames $\downarrow$ & Acc $\uparrow$ & \#Frames $\downarrow$ & Acc $\uparrow$ & \#Frames $\downarrow$ & Acc $\uparrow$ & \#Frames $\downarrow$ & Acc $\uparrow$ & \#Frames $\downarrow$ & Acc $\uparrow$ \\
\midrule

\multicolumn{11}{l}{\textit{Large Multimodal Models (Reference)}} \\
\midrule
GPT-4o~ & 384 & 71.9 & 32 & 58.2 & 60 & 48.9 & 32 & 42.0 & 0.5fps & 64.6 \\
GPT-5~ & 50 & 79.8 & 384 & 64.5 & 384 & 60.1 & 50 & 57.0 & 50 & 76.8 \\
Gemini-3.1-Flash-Lite~ & 128 & 71.9 & 128 & 66.7 & 128 & 54.2 & 128 & 49.8 & 128 & 74.0 \\
Qwen2.5-VL-72B~ & 768 & 73.3 & 768 & 60.7 & 768 & 47.3 & - & 40.2 & 768 & 74.6 \\

\midrule

\multicolumn{11}{l}{\textit{Video Agents on Larger or Proprietary Backbones (Reference)}} \\
\midrule
VideoSeek~ \textcolor{gray}{\scriptsize(GPT-5)} & 60.9 & 70.1 & 85.6 & 67.3 & 148.3 & 68.4 & 75.5 & 48.6 & - & - \\
VideoSeek~ \textcolor{gray}{\scriptsize(Qwen2.5-VL-72B)} & 223.8 & 51.9 & 295 & 36.7 & 245.9 & 34.2 & 210.6 & 33.7 & - & - \\
VideoAgent~ \textcolor{gray}{\scriptsize(GPT-4)} & 87.0 & 56.0 & - & - & 25.5 & 29.3 & - & - & - & - \\
\midrule

\multicolumn{11}{l}{\textit{Comparable Methods ($\sim$7--8B Open-Source Foundations)}} \\
\midrule
Qwen3-VL-8B-Instruct~ & 32 & 59.4 & 32 & \underline{58.0} & 32 & 38.3 & 32 & 42.7 & 32 & 61.3 \\
VideoChat-R1 & 192 & 60.0 & 32 & 49.1 & 32 & 34.3 & 32 & 33.0 & 32 & 54.3 \\
Video-R1~ & 32 & 59.9  & 32 & 49.1 & 32 & 35.3 & 32 & 36.5 & 32 & 60.2 \\
Video-MTR & 32 & 59.0 & 32 & 52.3 & 32 & 38.2 & - & - & 32 & 48.4 \\
Conan & 32 & \underline{60.5} & 32 & 56.6 & 32 & 39.2 & 32 & 44.6 & 32 & 63.4 \\
EVA~& 78.6 & 60.2 & 74.8 & 55.0 & 137.1 & \textbf{43.3} & 55.6 & 37.2 & 69 & \textbf{68.3} \\
FrameThinker~ & 24.1 & 47.6  & 21.1 & 52.9 & 23.9 & 36.6 & 15.9 & \underline{46.8} & 23.2 & 59.1 \\

\midrule
    
\textbf{ForeSeek-Base} & 44.2 & 54.1 & 46.9 & 52.2 & 62.3 & 38.7 & 41.4 & 34.0 & 41.8 & 50.7\\
\textbf{ForeSeek-SFT} & 45.6 & 56.5 & 42.0 & 49.6 & 52.1 & 37.8 & 39.9 & 43.4 & 49.5 & 57.9\\
\textbf{ForeSeek-GSPO} & 24.8 & \textbf{61.1} & 24.1 & \textbf{60.1} & 31.0 & \underline{40.3} & 22.0 & \textbf{53.2} & 23.4 & \underline{66.2}\\
\bottomrule
\end{tabular}%
}
\caption{Performance comparison on five video understanding benchmarks.
Accuracy and the number of consumed frames are reported for each benchmark.
Methods are grouped by backbone setting, with large or proprietary models listed as references. 
\textbf{Bold} and \underline{underline} denote the best and second-best accuracy among all entries in the comparable $\sim$7--8B open-source group.}
\label{tab:main_results}
\end{table*}

\begin{table}[t]
\centering
\setlength{\tabcolsep}{4pt}   

\begin{tabular}{l cc cc}
\toprule
\multirow{2}{*}{\textbf{Benchmark}} &
\multicolumn{2}{c}{\textbf{Unif. 32f}} &
\multicolumn{2}{c}{\textbf{Agent (ours)}} \\
\cmidrule(lr){2-3}\cmidrule(lr){4-5}
& All & V-Nat. & All & V-Nat. \\
\midrule
VideoMME       & 59.4 & 26.0 & 61.1\,\textcolor{gray}{\scriptsize(+1.7)} & 30.6\,\textcolor{gray}{\scriptsize\textbf{(+4.6)}}
\\
LongVideoBench & 58.0 & 31.6 & 60.1\,\textcolor{gray}{\scriptsize(+2.1)} & 38.8\,\textcolor{gray}{\scriptsize\textbf{(+7.2)}}
\\
LVBench        & 38.3 & 21.7 & 40.3\,\textcolor{gray}{\scriptsize(+2.0)} & 25.2\,\textcolor{gray}{\scriptsize\textbf{(+3.5)}}
\\
VideoHolmes   & 42.7 & 22.6 & 53.2\,\textcolor{gray}{\scriptsize(+10.5)} & 40.2\,\textcolor{gray}{\scriptsize\textbf{(+17.6)}}
\\
\bottomrule
\end{tabular}
\caption{Accuracy on the Video-Oasis~\cite{videooasis} \emph{video-native} (V-Nat.) subset vs.\ the full set. MLVU is omitted as Video-Oasis does not provide annotations for it.}
\label{tab:video_native}
\end{table}

\subsection{Main Results}
\label{sec:main_results}
Table~\ref{tab:main_results} reports accuracy (Acc$\uparrow$) and frames consumed (\#Frames$\downarrow$, counting underlying image frames regardless of resolution or packing) across five benchmarks, focusing on comparable $\sim$7--8B open-source agents with larger or closed-source models as references.

\paragraph{Efficiency--accuracy trade-off.}
\our{}-GSPO occupies a favorable region of the frame--accuracy curve, using $2.5$--$4.4\times$ fewer frames than EVA ($22.0$--$31.0$ vs.\
$55.6$--$137.1$) while matching or exceeding it on three of five benchmarks
and trailing by at most $3$ points on the others. Against our uniform 32-frame Qwen3-VL-8B reference, it gains
$+1.7/+2.1/+2.0/+10.5/+4.9$ on VideoMME, LongVideoBench, LVBench, VideoHolmes, and MLVU, respectively. Although using up to $38\%$ more frames than FrameThinker, \our{}-GSPO outperforms it on every benchmark. It also surpasses Conan at a similar frame budget, including $+8.6$ on VideoHolmes with only $22.0$ frames.

\paragraph{Generalization to unseen distributions.}
Specifically, our SFT/RL corpus draws from CG-Bench, LLaVA-Video, LongVideoBench, and VideoHolmes, so the latter two are in-distribution (ID), while VideoMME, LVBench, and MLVU contribute no training data and are out-of-distribution (OOD).
Note that the per-round cap in Analysis~(a) is swept on VideoMME, LongVideoBench, and LVBench, so for those we claim generalization of the policy rather than of the inference configuration; the deployed cap is a single efficiency-oriented value, and MLVU is untouched by both.
Comparable baselines train on non-overlapping corpora (Conan-91K, LongVideo-Reason, Video-R1-CoT-165k, NExT-GQA/QVHighlights), so the same three benchmarks are OOD for them too.
On the OOD triple, \our{}-GSPO leads on VideoMME ($61.1$, $+0.6$ over Conan) and stays within a few points of EVA on LVBench ($40.3$, $-3.0$) and MLVU ($66.2$, $-2.1$) at a fraction of its frame budget.
On the ID pair it leads clearly (LongVideoBench $60.1$, $+3.5$ over Conan; VideoHolmes $53.2$, $+6.4$ over FrameThinker).
This ID lead is not an exposure effect: VideoHolmes contributes only $3.0\%$ of the SFT corpus and $6.7\%$ of the RL prompts, yet yields our largest margin ($+10.5$ over the uniform $32$-frame backbone), whereas LongVideoBench---at $28.2\%$---gains only $+2.1$, and the zero-exposure MLVU gains $+4.9$.
Gains therefore track reasoning demand rather than in-distribution exposure.

\paragraph{Effect of two-stage training.}
\our{}-Base runs the full \our{} loop on the untrained backbone by prompting alone.
Under fixed backbone, teacher, and data, the Base $\rightarrow$ SFT $\rightarrow$ GSPO progression reveals a division of labor. SFT injects the foresight schema, sharply lifting reasoning-heavy benchmarks (VideoHolmes $+9.4$, MLVU $+7.2$, VideoMME $+2.4$) but slightly regressing on LongVideoBench ($-2.6$) and LVBench ($-0.9$)---typical of imitation learning that rewards trajectory conformance over task accuracy. GSPO corrects this with non-negative gains everywhere (VideoMME $+4.6$, LongVideoBench $+10.5$, LVBench $+2.5$, VideoHolmes $+9.8$, MLVU $+8.3$) while cutting frames on all five, since its length penalty activates only when the answer is correct. End-to-end, \our{}-GSPO surpasses \our{}-Base by $+1.6$ to $+19.2$ points at $44$--$50\%$ lower frame cost, 
and on VideoHolmes it exceeds GPT-4o ($+11.2$) despite a smaller backbone; 
residual gaps on LVBench (vs.\ GPT-4o $48.9$ / GPT-5 $60.1$) and MLVU (vs.\ EVA $68.3$) suggest that active seeking \emph{partially} substitutes for---but does not fully replace---raw backbone capacity.

\begin{figure*}[!t]
\centering
\includegraphics[width=0.90\textwidth]{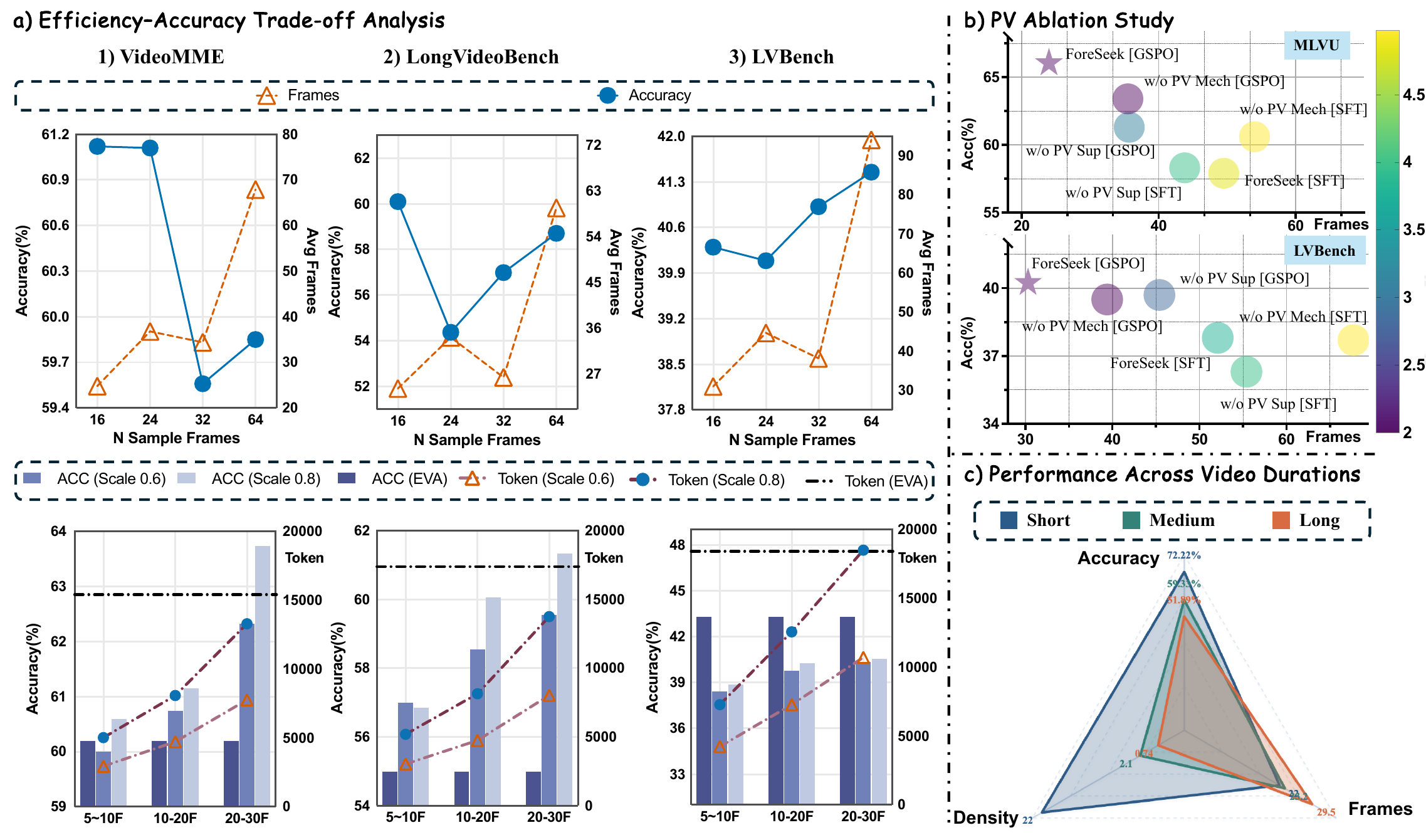}
\caption{
\textbf{Unified analysis of \our.}
(a, top) Per-round frame-cap sweep on VideoMME, LongVideoBench, and LVBench;
(a, bottom) Phase-1 initial-budget and compression-scale trade-offs, with EVA
as a fixed-cost reference;
(b) ablation of the prediction--verification mechanism
(\textit{\our{}}, \textit{w/o PV Supervision}, and \textit{w/o PV Mechanism})
under SFT and RL;
(c) sub-linear budget scaling across VideoMME duration buckets.
}
\label{fig:efficiency_analysis}
\end{figure*}

\subsection{Evaluation on Video-Native Questions}
\label{sec:video_native}
The high aggregate accuracy of Qwen3-VL-8B-Instruct under uniform $32$-frame sampling is partly inflated by \emph{shortcut-solvable} questions answerable from linguistic priors, a single frame, or audio cues, making the overall gains in Table~\ref{tab:video_native} ($+1.7$ to $+10.5$) a conservative measure of our benefit.
Following Video-Oasis~\cite{videooasis}, we isolate each benchmark's \emph{video-native} subset---questions that strictly require visual--temporal reasoning---and compare our agent (adaptive $22$--$31$ frames) with the same baseline under identical per-frame token budgets, so that \emph{only the allocation strategy differs}.
On this subset, gains rise to $+4.6$/$+7.2$/$+3.5$/$+17.6$ on VideoMME /
LongVideoBench / LVBench / VideoHolmes, respectively, showing that
\our{}'s benefit is concentrated where visual--temporal evidence must be sought.

\subsection{Analysis and Ablations}
\label{sec:analysis}
Figure~\ref{fig:efficiency_analysis} summarizes four analyses in three groups: budget studies (a), prediction--verification ablation (b), and per-length-bucket analysis (c).

\paragraph{(a) Budget studies.}
Sweeping the per-round frame cap over $\{16,24,32,64\}$ (top) yields benchmark-dependent, non-monotone effects: VideoMME and LongVideoBench peak at $16$ frames ($61.14$/$60.06$), whereas LVBench favors $64$ ($41.45$).
Since larger caps provide no consistent gain, we set the efficiency-oriented default to $n_r\in[4,16]$.
Expanding the Phase-1 range from $5$--$10$ to $20$--$30$ frames (bottom) consistently improves all three benchmarks and compression scales: VideoMME rises from $60.00$ to $62.33$ at $s{=}0.6$ and from $60.59$ to $63.74$ at $s{=}0.8$, while LVBench rises from $38.41$ to $40.35$ and from $38.86$ to $40.54$. By contrast, increasing $s$ from $0.60$ to $0.80$ adds only about one point at $1.7\times$ the token cost, indicating that tokens are better spent on more frames than higher per-frame resolution. 
At $s{=}0.6$ we consume only $19$--$58\%$ of EVA's tokens, which is enough to surpass it on VideoMME once the Phase-1 range reaches $20$--$30$ frames ($62.33$ vs.\ $60.2$), whereas on LVBench a gap of about $3$ points remains.
For the headline results, we retain the duration-adaptive default in Eq.~\ref{eq:phase1_sampling} ($10$--$20$ frames, $s{=}0.8$) to remain within the frame budgets of efficient baselines and use one benchmark-agnostic configuration across all five benchmarks. When tokens rather than frames are the primary constraint, we recommend $20$--$30$ frames with $s{=}0.6$, which is more token-efficient and raises VideoMME to $62.33$.


\paragraph{(b) Prediction--verification ablation.}
We compare the full \textit{ForeSeek} trajectory with \textit{w/o PV Supervision}, which removes explicit prediction--verification supervision, and \textit{w/o PV Mechanism}, which further removes the reasoning process.
The colored bars on the right indicate the average total number of reasoning rounds.
After SFT, the effects are mixed across LVBench and MLVU, with no consistent advantage from the complete structure. 
After RL, however, ForeSeek performs best on both benchmarks, outperforming the two ablations by $+0.6$/$+0.8$ points on LVBench and $+4.9$/$+2.8$ points on MLVU. It also uses fewer frames and rounds: $31.0$ frames and $2.00$ rounds vs.\ $45.4$/$39.4$ and $3.26$/$2.60$ on LVBench and $23.4$ frames and $2.00$ rounds vs.\ $35.7$/$35.5$ frames and $3.07$/$2.16$ rounds on MLVU. These results indicate that RL is necessary for prediction--verification to yield consistent accuracy--efficiency gains.

\paragraph{(c) Analysis across video-length buckets.}
Panel (c) contrasts accuracy, total frames, and sampling density across VideoMME's short/medium/long buckets ($1$/$11$/$40$\, min). As duration grows $40\times$, total frames rise only $1.34\times$ ($22.0{\to}23.2{\to}29.5$), so density collapses $29.7\times$ ($22.0{\to}2.1{\to}0.74$\,f/min): the agent commits a nearly constant per-question budget instead of scaling it with video length. 
Accuracy declines with video length ($72.22{\to}59.33{\to}51.89$), while
correct and incorrect predictions use similar budgets, indicating that
sampling does not increase with uncertainty.

\section{Conclusion}


We present \textbf{\our}, a foresight-driven video agent that reframes long-video reasoning as predictive exploration rather than reactive retrieval, maintaining visually grounded expectations over unseen regions that are refined by both confirming and contradicting evidence.
Trained with trajectory supervision and reinforcement learning, it attains a strong accuracy--efficiency trade-off on five long-video benchmarks with far fewer frames.
We hope this encourages future agents to anticipate unseen content rather than merely retrieve it.

\bibliography{aaai2027}

\setcounter{secnumdepth}{2}
\appendix
\renewcommand{\topfraction}{0.95}
\renewcommand{\bottomfraction}{0.90}
\renewcommand{\dbltopfraction}{0.95}
\renewcommand{\textfraction}{0.05}
\renewcommand{\floatpagefraction}{0.85}
\renewcommand{\dblfloatpagefraction}{0.85}
\setcounter{topnumber}{4}
\setcounter{bottomnumber}{4}
\setcounter{totalnumber}{6}
\setcounter{dbltopnumber}{3}

\twocolumn[{%
\begin{@twocolumnfalse}
\section{Appendix}
\label{sec:appendix}
\vspace{0.4em}
\centering
\resizebox{\textwidth}{!}{%
\begin{tabular}{l c c c c c c c c c c}
\toprule
\multirow{2}{*}{\textbf{Method (Qwen2.5-VL-7B)}} &
\multicolumn{2}{c}{\textbf{VideoMME (w/o sub)}} &
\multicolumn{2}{c}{\textbf{LongVideoBench}} &
\multicolumn{2}{c}{\textbf{LVBench (w/o sub)}} &
\multicolumn{2}{c}{\textbf{VideoHolmes}} &
\multicolumn{2}{c}{\textbf{MLVU (w/o sub)}} \\
\cmidrule(lr){2-3} \cmidrule(lr){4-5} \cmidrule(lr){6-7} \cmidrule(lr){8-9} \cmidrule(lr){10-11}
& \#Frames $\downarrow$ & Acc $\uparrow$ & \#Frames $\downarrow$ & Acc $\uparrow$ & \#Frames $\downarrow$ & Acc $\uparrow$ & \#Frames $\downarrow$ & Acc $\uparrow$ & \#Frames $\downarrow$ & Acc $\uparrow$ \\
\midrule
VideoChat-R1   & 192  & 60.0 & 32   & 49.1 & 32    & 34.3 & 32   & 33.0 & 32   & 54.3 \\
Video-R1       & 32   & 59.9 & 32   & 49.1 & 32    & 35.3 & 32   & 36.5 & 32   & 60.2 \\
Video-MTR      & 32   & 59.0 & 32   & 52.3 & 32    & 38.2 & --   & --   & 32   & 48.4 \\
Conan          & 32   & \textbf{60.5} & 32   & \underline{56.6} & 32    & 39.2 & 32   & 44.6 & 32   & \underline{63.4} \\
EVA            & 78.6 & \underline{60.2} & 74.8 & 55.0 & 137.1 & \textbf{43.3} & 55.6 & 37.2 & 69.0 & \textbf{68.3} \\
FrameThinker   & 24.1 & 47.6 & 21.1 & 52.9 & 23.9  & 36.6 & 15.9 & \underline{46.8} & 23.2 & 59.1 \\
\midrule
\textbf{\our{}-Base} & 62.3 & 43.9 & 59.5 & 46.5 & 82.5  & 33.3 & 52.7 & 27.9 & 62.7 & 47.5 \\
\textbf{\our{}-SFT}  & 44.9 & 50.9 & 41.8 & 46.7 & 54.6  & 35.0 & 38.8 & 36.7 & 44.4 & 53.4 \\
\textbf{\our{}-GSPO} & 24.8 & 59.4 & 24.0 & \textbf{59.1} & 31.0  & \underline{40.4} & 22.0 & \textbf{47.1} & 23.4 & 62.6 \\
\bottomrule
\end{tabular}%
}
\captionof{table}{Same-backbone comparison on \textbf{Qwen2.5-VL-7B} across all five benchmarks. \textbf{Bold} and \underline{underline} denote the best and second-best accuracy.}
\label{tab:qwen25_main}
\vspace{1.0em}
\end{@twocolumnfalse}
}]

\subsection{Same-Backbone Comparison on Qwen2.5-VL-7B}
\label{app:qwen25}

Most open-source video agents in Table~\ref{tab:main_results} are built upon \textbf{Qwen2.5-VL-7B}, whereas our main experiments adopt \textbf{Qwen3-VL-8B}. To decouple the contribution of \our{} from backbone scaling, we train and evaluate \our{} directly on Qwen2.5-VL-7B under the identical two-stage pipeline (SFT followed by GSPO).
As shown in Table~\ref{tab:qwen25_main}, under the same Qwen2.5-VL-7B backbone, \our{}-GSPO achieves the best accuracy on LongVideoBench ($59.1\%$) and VideoHolmes ($47.1\%$), surpasses all 32-frame 7B agents on LVBench ($40.4\%$), and rivals EVA on VideoMME ($59.4\%$ at $24.8$ frames vs.\ $60.2\%$ at $78.6$ frames) with $3$--$4\times$ fewer frames across benchmarks.

\subsection{Extended Prediction--Verification Ablations}
\label{app:pv}

In Section~\ref{sec:main_results} (Figure~\ref{fig:efficiency_analysis}b), we reported the prediction--verification (PV) ablation on LVBench and MLVU. Table~\ref{tab:pv_app} extends this strictly matched post-RL comparison to four benchmarks: VideoMME, LongVideoBench, LVBench, and VideoHolmes.

\begin{table}[!htbp]
\centering
\small
\setlength{\tabcolsep}{2.5pt}
\resizebox{\columnwidth}{!}{%
\begin{tabular}{l cc cc cc cc}
\toprule
\multirow{2}{*}{\textbf{Variant}} &
\multicolumn{2}{c}{\textbf{VideoMME}} &
\multicolumn{2}{c}{\textbf{LVB}} &
\multicolumn{2}{c}{\textbf{LVBench}} &
\multicolumn{2}{c}{\textbf{Holmes}} \\
\cmidrule(lr){2-3} \cmidrule(lr){4-5} \cmidrule(lr){6-7} \cmidrule(lr){8-9}
& Fr.$\downarrow$ & Acc$\uparrow$ & Fr.$\downarrow$ & Acc$\uparrow$ & Fr.$\downarrow$ & Acc$\uparrow$ & Fr.$\downarrow$ & Acc$\uparrow$ \\
\midrule
\textbf{\our{} (Full)} & \textbf{24.8} & \textbf{61.1} & \textbf{24.1} & \textbf{60.1} & \textbf{31.0} & \textbf{40.3} & \textbf{22.0} & \textbf{53.2} \\
w/o PV Sup             & 36.3 & 59.7 & 39.2 & 55.1 & 45.4 & 39.7 & 34.8 & 51.3 \\
w/o PV Mech            & 27.6 & 58.1 & 29.2 & 57.1 & 39.4 & 39.5 & 24.2 & 52.2 \\
\bottomrule
\end{tabular}%
}
\caption{Strictly matched post-RL ablations of the prediction--verification (PV) mechanism across four benchmarks.}
\label{tab:pv_app}
\end{table}

\paragraph{Experimental Control and Negative Evidence.}
All variants share identical training data, optimization steps, frame caps, and rewards. \emph{w/o PV Sup} keeps the same data construction but strips PV text from training trajectories and PV modules from the agent; \emph{w/o PV Mech} drops PV from both trajectory synthesis and inference from scratch, representing a standard adaptive-search agent trained with the identical SFT+GSPO pipeline. \our{} achieves the highest accuracy ($+0.8$ to $+3.0$ points over \emph{w/o PV Mech}) with the fewest frames on all four benchmarks. While demoting unverified intervals (\texttt{not\_found}) under sparse sampling carries an inherent efficiency--recall trade-off, rejecting non-supporting intervals filters out irrelevant visual context and yields a clear net accuracy gain over the $\sim$35--45-frame \emph{w/o PV Sup} baseline.

\subsection{Temporal Localization Evaluation}
\label{app:loc}

To verify whether foresight-guided search localizes the true evidentiary segments, we evaluate the temporal overlap between the agent's final inspected interval and ground-truth evidence spans across all five benchmarks. Table~\ref{tab:loc} reports mean IoU (mIoU) and Recall at $\text{IoU} \ge 0.5$ ($\text{R@0.5}$) against EVA.

\begin{table}[!htbp]
\centering
\small
\setlength{\tabcolsep}{5pt}
\begin{tabular}{l cc cc}
\toprule
\multirow{2}{*}{\textbf{Benchmark}} &
\multicolumn{2}{c}{\textbf{mIoU (\%)} $\uparrow$} &
\multicolumn{2}{c}{\textbf{R@0.5 (\%)} $\uparrow$} \\
\cmidrule(lr){2-3} \cmidrule(lr){4-5}
& \textbf{\our{}} & EVA & \textbf{\our{}} & EVA \\
\midrule
VideoMME       & \textbf{24.1} & 19.5 & \textbf{16.2} & 8.1 \\
LongVideoBench & \textbf{13.9} & 10.8 & \textbf{8.6}  & 3.7 \\
LVBench        & \textbf{4.4}  & 3.2  & \textbf{1.5}  & 0.8 \\
VideoHolmes    & 27.3 & \textbf{32.9} & 17.6 & \textbf{23.9} \\
MLVU           & \textbf{22.4} & 17.1 & \textbf{16.4} & 6.0 \\
\bottomrule
\end{tabular}
\caption{Temporal localization comparison between \our{} and EVA in terms of mIoU and $\text{R@0.5}$.}
\label{tab:loc}
\end{table}

Despite observing $2.5$--$4.4\times$ fewer frames, \our{} outperforms EVA in both mIoU and $\text{R@0.5}$ on four of the five benchmarks, nearly doubling $\text{R@0.5}$ on VideoMME ($16.2\%$ vs.\ $8.1\%$) and more than doubling it on LongVideoBench ($8.6\%$ vs.\ $3.7\%$) and MLVU ($16.4\%$ vs.\ $6.0\%$).

\subsection{Inference Efficiency and Protocol Details}
\label{app:cost}

\paragraph{VLM Calls, Tokens, and Deterministic Decoding.}
Table~\ref{tab:cost} evaluates \our{} under deterministic greedy decoding and compares its accuracy, average VLM calls per question, cumulative frames, and total (visual + text) tokens against EVA.

\begin{table}[!htbp]
\centering
\small
\setlength{\tabcolsep}{2.2pt}
\resizebox{\columnwidth}{!}{%
\begin{tabular}{l cc cc cc cc}
\toprule
\multirow{2}{*}{\textbf{Benchmark}} &
\multicolumn{2}{c}{\textbf{Acc} $\uparrow$} &
\multicolumn{2}{c}{\textbf{Calls} $\downarrow$} &
\multicolumn{2}{c}{\textbf{Frames} $\downarrow$} &
\multicolumn{2}{c}{\textbf{Tokens} $\downarrow$} \\
\cmidrule(lr){2-3} \cmidrule(lr){4-5} \cmidrule(lr){6-7} \cmidrule(lr){8-9}
& EVA & \textbf{Ours} & EVA & \textbf{Ours} & EVA & \textbf{Ours} & EVA & \textbf{Ours} \\
\midrule
VideoMME        & 60.2 & \textbf{61.2} & 2.41 & \textbf{2.01} & 82.0  & \textbf{25.1} & 16.9k & \textbf{10.5k} \\
LongVideoBench  & 55.0 & \textbf{58.9} & 2.57 & \textbf{2.00} & 85.3  & \textbf{24.1} & 19.0k & \textbf{10.6k} \\
LVBench         & \textbf{43.3} & 41.0 & 2.63 & \textbf{2.01} & 133.6 & \textbf{31.1} & 20.5k & \textbf{15.5k} \\
VideoHolmes     & 37.2 & \textbf{52.9} & 2.75 & \textbf{2.02} & 82.7  & \textbf{22.3} & 9.1k  & \textbf{5.0k}  \\
MLVU            & \textbf{68.3} & 66.5 & 2.42 & \textbf{2.01} & 93.7  & \textbf{23.5} & 16.6k & \textbf{11.8k} \\
\bottomrule
\end{tabular}%
}
\caption{End-to-end efficiency comparison between EVA and deterministic \our{} (greedy decoding).}
\label{tab:cost}
\end{table}

Because visual tokens dominate multimodal context while structured prediction text adds minimal overhead, \our{} reduces VLM calls from $2.41$--$2.75$ to $\sim 2.01$ and cuts total tokens by $25\%$--$45\%$. Deterministic greedy decoding also closely matches Table~\ref{tab:main_results}, confirming stability.

\paragraph{Dataset Split Isolation.}
We enforce zero overlap at both the video (\texttt{video\_uid}) and question (normalized text) levels between training and evaluation: LongVideoBench training uses the unlabeled test split while evaluation uses the validation split; VideoHolmes training and evaluation use the official disjoint train and test splits; VideoMME, LVBench, and MLVU are strictly out-of-distribution.

\end{document}